\documentclass[runningheads]{llncs}
\usepackage{graphicx}
\usepackage{booktabs}
\usepackage{multirow}
\usepackage{subfig}
\usepackage{amsmath}
\usepackage{amssymb}

\begin{document}
%
%\title{Contribution Title\thanks{Supported by organization x.}}
%\title{Passenger-Vehicle~Interactions in Autonomous~Vehicles: Intent~Recognition and Slot~Extraction from Spoken Utterances}
\title{Natural~Language~Interactions in Autonomous~Vehicles: Intent~Detection and Slot~Filling from Passenger Utterances}
%\titlerunning{Passenger-Vehicle~Interactions in Autonomous~Vehicles}
%\titlerunning{NLI in AVs: Intent Detection and Slot Filling from Passenger Utterances}
\titlerunning{Natural~Language~Interactions in AVs: Intent Detection and Slot Filling}
%\author{First Author\inst{1}\orcidID{0000-1111-2222-3333} \and
%Second Author\inst{2,3}\orcidID{1111-2222-3333-4444} \and
%Third Author\inst{3}\orcidID{2222--3333-4444-5555}}
%\author{Anonymous Author(s)}
\author{Eda~Okur \and
Shachi~H~Kumar \and
Saurav~Sahay \and 
Asli~Arslan~Esme \and 
Lama~Nachman}
%\authorrunning{Anonymous Author(s) et al.}
\authorrunning{E. Okur et al.}
%
%\institute{Princeton University, Princeton NJ 08544, USA \and
%Springer Heidelberg, Tiergartenstr. 17, 69121 Heidelberg, Germany
%\institute{Affiliation, Address\\
\institute{Intel Labs, Anticipatory Computing Lab, USA\\
\email{\{eda.okur, shachi.h.kumar, saurav.sahay, asli.arslan.esme, lama.nachman\}@intel.com}}
\maketitle              % typeset the header of the contribution
\begin{abstract}
 Understanding passenger intents and extracting relevant slots are important building blocks towards developing contextual dialogue systems for natural interactions in autonomous vehicles (AV). In this work, we explored AMIE (Automated-vehicle Multi-modal In-cabin Experience), the in-cabin agent responsible for handling certain passenger-vehicle interactions. When the passengers give instructions to AMIE, the agent should parse such commands properly and trigger the appropriate functionality of the AV system. In our current explorations, we focused on AMIE scenarios describing usages around setting or changing the destination and route, updating driving behavior or speed, finishing the trip and other use-cases to support various natural commands. We collected a multi-modal in-cabin dataset with multi-turn dialogues between the passengers and AMIE using a Wizard-of-Oz scheme via a realistic scavenger hunt game activity. After exploring various recent Recurrent Neural Networks (RNN) based techniques, we introduced our own hierarchical joint models to recognize passenger intents along with relevant slots associated with the action to be performed in AV scenarios. Our experimental results outperformed certain competitive baselines and achieved overall F1-scores of 0.91 for utterance-level intent detection and 0.96 for slot filling tasks. In addition, we conducted initial speech-to-text explorations by comparing intent/slot models trained and tested on human transcriptions versus noisy Automatic Speech Recognition (ASR) outputs. Finally, we compared the results with single passenger rides versus the rides with multiple passengers.

%\keywords{First keyword  \and Second keyword \and Another keyword.}
\keywords{Intent recognition  \and Slot filling \and Hierarchical joint learning \and Spoken language understanding (SLU) \and In-cabin dialogue agent.}
\end{abstract}
\section{Introduction}

One of the exciting yet challenging areas of research in Intelligent Transportation Systems is developing context-awareness technologies that can enable autonomous vehicles to interact with their passengers, understand passenger context and situations, and take appropriate actions accordingly. To this end, building multi-modal dialogue understanding capabilities situated in the in-cabin context is crucial to enhance passenger comfort and gain user confidence in AV interaction systems. Among many components of such systems, intent recognition and slot filling modules are one of the core building blocks towards carrying out successful dialogue with passengers. As an initial attempt to tackle some of those challenges, this study introduce in-cabin intent detection and slot filling models to identify passengers' intent and extract semantic frames from the natural language utterances in AV. The proposed models are developed by leveraging User Experience (UX) grounded realistic (ecologically valid) in-cabin dataset. This dataset is generated with naturalistic passenger behaviors, multiple passenger interactions, and with presence of a Wizard-of-Oz (WoZ) agent in moving vehicles with noisy road conditions.

\subsection{Background}

Long Short-Term Memory (LSTM) networks \cite{lstm-1997} are widely-used for temporal sequence learning or time-series modeling in Natural Language Processing (NLP). These neural networks are commonly employed for sequence-to-sequence (seq2seq) and sequence-to-one (seq2one) modeling problems, including slot filling tasks \cite{mesnil-2015} and utterance-level intent classification \cite{hakkani-2016,lstm-2015} which are well-studied for various application domains. Bidirectional LSTMs (Bi-LSTMs) \cite{bi-lstm-1997} are extensions of traditional LSTMs which are proposed to improve model performance on sequence classification problems even further. Jointly modeling slot extraction and intent recognition \cite{hakkani-2016,zhang-2016} is also explored in several architectures for task-specific applications in NLP. Using Attention mechanism \cite{attention-2015,attention-2016} on top of RNNs is yet another recent break-through to elevate the model performance by attending inherently crucial sub-modules of given input. There exist various architectures to build hierarchical learning models \cite{h-zhou-2016,h-meng-2017,wen-2018} for document-to-sentence level, and sentence-to-word level classification tasks, which are highly domain-dependent and task-specific.

Automatic Speech Recognition (ASR) technology has recently achieved human-level accuracy in many fields \cite{ASR-2016,ASR-2017}. For spoken language understanding (SLU), it is shown that training SLU models on true text input (i.e., human transcriptions) versus noisy speech input (i.e., ASR outputs) can achieve varying results \cite{asr-liu-2016}. Even greater performance degradations are expected in more challenging and realistic setups with noisy environments, such as moving vehicles in actual traffic conditions. As an example, a recent work \cite{Zheng-2017} attempts to classify sentences as navigation-related or not using the DARPA supported CU-Move in-vehicle speech corpus \cite{CU-Move-2001}, a relatively old and large corpus focusing on route navigation. For this binary intent classification task, the authors observed that detection performances are largely affected by high ASR error rates due to background noise and multi-speakers in CU-Move dataset (not publicly available). For in-cabin dialogue between car assistants and driver/passengers, recent studies explored creating a public dataset using a WoZ approach \cite{stanford-2017}, and improving ASR for passenger speech recognition \cite{asr-fukui-2018}.

A preliminary report on research designed to collect data for human-agent interactions in a moving vehicle is presented in a previous study \cite{HRI-2018}, with qualitative analysis on initial observations and user interviews. Our current study is focused on the quantitative analysis of natural language interactions found in this in-vehicle dataset \cite{WiML-2018}, where we address intent detection and slot extraction tasks for passengers interacting with the AMIE in-cabin agent.

%\paragraph{Contributions}
\subsubsection{Contributions.}
%We introduced AMIE in-cabin interaction agent, 
In this study, we propose a UX grounded realistic intent recognition and slot filling models with naturalistic passenger-vehicle interactions in moving vehicles. Based on observed interactions, we defined in-vehicle intent types and refined their relevant slots through a data driven process. After exploring existing approaches for jointly training intents and slots, we applied certain variations of these models that perform best on our dataset to support various natural commands for interacting with the car-agent. The main differences in our proposed models can be summarized as follows: (1) Using the extracted intent keywords in addition to the slots to jointly model them with utterance-level intents (where most of the previous work \cite{h-zhou-2016,h-meng-2017} only join slots and utterance-level intents, ignoring the intent keywords); (2) The 2-level hierarchy we defined by word-level detection/extraction for slots and intent keywords first, and then filtering-out predicted non-slot and non-intent keywords instead of feeding them into the upper levels of the network (i.e., instead of using stacked RNNs with multiple recurrent hidden layers for the full utterance \cite{h-meng-2017,wen-2018}, which are computationally costly for long utterances with many non-slot \& non-intent-related words), and finally using only the predicted valid-slots and intent-related keywords as an input to the second level of the hierarchy; (3) Extending joint models \cite{hakkani-2016,zhang-2016} to include both beginning-of-utterance and end-of-utterance tokens to leverage Bi-LSTMs (after observing that we achieved better results by doing so). We compared our intent detection and slot filling results with the results obtained from Dialogflow\footnote{https://dialogflow.com}, a commercially available intent-based dialogue system by Google, and showed that our proposed models perform better for both tasks on the same dataset. We also conducted initial speech-to-text explorations by comparing models trained and tested (10-fold CV) on human transcriptions versus noisy ASR outputs (via Cloud Speech-to-Text\footnote{https://cloud.google.com/speech-to-text/}). Finally, we compared the results with single passenger rides versus the rides with multiple passengers.

\section{Methodology}
\label{method}

\subsection{Data Collection and Annotation}

\begin{figure}[t]
\includegraphics[width=\textwidth]{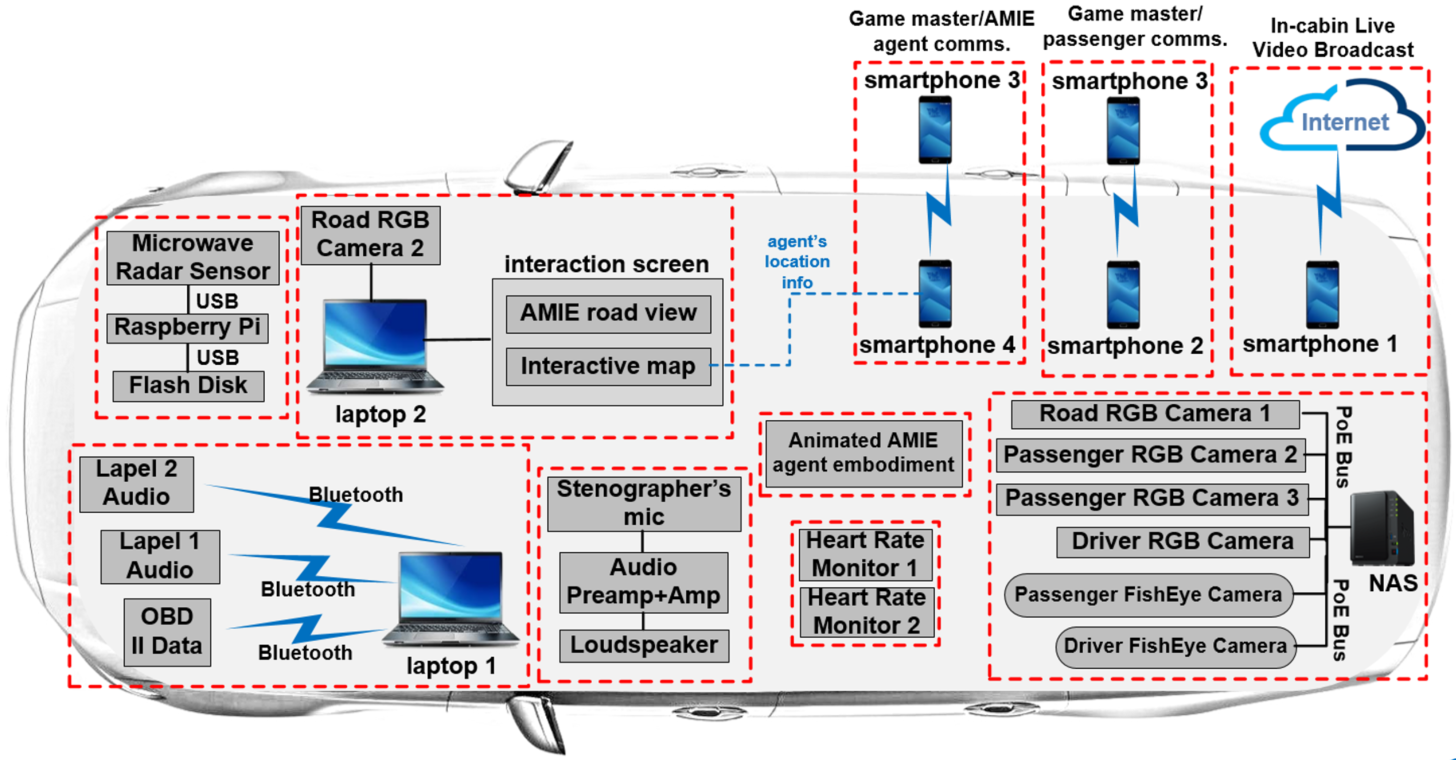}
\caption{AMIE In-cabin Data Collection Setup} \label{fig:car}
\end{figure}

Our AV in-cabin dataset includes around 30 hours of multi-modal data collected from 30 passengers (15 female, 15 male) in a total of 20 rides/sessions. In 10 sessions, single passenger was present (i.e., \textit{singletons}), whereas the remaining 10 sessions include two passengers (i.e., \textit{dyads}) interacting with the vehicle. The data is collected "in the wild" on the streets of Richmond, British Columbia, Canada. Each ride lasted about 1 hour or more. The vehicle is modified to hide the operator and the human acting as in-cabin agent from the passengers, using a variation of WoZ approach \cite{WoZ-2017}. Participants sit in the back of the car and are separated by a semi-sound proof and translucent screen from the human driver and the WoZ AMIE agent at the front. In each session, the participants were playing a scavenger hunt game by receiving instructions over the phone from the Game Master. Passengers treat the car as AV and communicate with the WoZ AMIE agent via speech commands. Game objectives require passengers to interact naturally with the agent to go to certain destinations, update routes, stop the vehicle, give specific directions regarding where to pull over or park (sometimes with gesture), find landmarks, change speed, get in and out of the vehicle, etc. Further details of the data collection design and scavenger hunt protocol can be found in the preliminary study \cite{HRI-2018}. See Fig.~\ref{fig:car} for the vehicle instrumentation to enhance multi-modal data collection setup. Our study is the initial work on this multi-modal dataset to develop intent detection and slot filling models, where we leveraged from the back-driver video/audio stream recorded by an RGB camera (facing towards the passengers) for manual transcription and annotation of in-cabin utterances. In addition, we used the audio data recorded by Lapel 1 Audio and Lapel 2 Audio (Fig.~\ref{fig:car}) as our input resources for the ASR.

For in-cabin intent understanding, we described 4 groups of usages to support various natural commands for interacting with the vehicle: (1) \textit{Set/Change Destination/Route} (including turn-by-turn instructions), (2) \textit{Set/Change Driving Behavior/Speed}, (3) \textit{Finishing the Trip Use-cases}, and (4) \textit{Others} (open/close door/window/trunk, turn music/radio on/off, change AC/temperature, show map, etc.). According to those scenarios, 10 types of passenger intents are identified and annotated as follows: \textit{SetDestination}, \textit{SetRoute}, \textit{GoFaster}, \textit{GoSlower}, \textit{Stop}, \textit{Park}, \textit{PullOver}, \textit{DropOff}, \textit{OpenDoor}, and \textit{Other}. For slot filling task, relevant slots are identified and annotated as: \textit{Location}, \textit{Position/Direction}, \textit{Object}, \textit{Time Guidance}, \textit{Person}, \textit{Gesture/Gaze} (e.g., \lq this\rq, \lq that\rq, \lq over there\rq, etc.), and \textit{None/O}. In addition to utterance-level intents and slots, word-level intent related keywords are annotated as \textit{Intent}. We obtained 1331 utterances having commands to AMIE agent from our in-cabin dataset. We expanded this dataset via the creation of similar tasks on Amazon Mechanical Turk \cite{amt-2012} and reached 3418 utterances with intents in total. Intent and slot annotations are obtained on the transcribed utterances by majority voting of 3 annotators. Those annotation results for utterance-level intent types, slots and intent keywords can be found in Table~\ref{D1} and Table~\ref{D2} as a summary of dataset statistics.

\addtolength{\tabcolsep}{+3pt} 
%\begin{table*}[t]
\begin{table}[t]
  \caption{AMIE Dataset Statistics: Utterance-level Intent Types}
  \label{D1}
  \centering
  %\resizebox{\textwidth}{!}{
  \begin{tabular}{*3c}
    \toprule
    \textbf{AMIE Scenario} & \textbf{Intent Type} & \textbf{Utterance Count} \\
    \toprule
     & Stop & 317 \\
    Finishing the Trip & Park & 450 \\
    Use-cases & PullOver & 295 \\
     & DropOff & 281 \\
    \midrule
    Set/Change & SetDestination & 552 \\
    Destination/Route & SetRoute & 676 \\
    \midrule
    Set/Change & GoFaster & 265 \\
    Driving Behavior/Speed & GoSlower & 238 \\
    \midrule
    Others & OpenDoor & 142 \\
    (Door, Music, etc.) & Other & 202 \\
    \midrule
     & \textit{Total} & \textit{3418} \\
    \bottomrule
  \end{tabular}
  %}
\end{table}
\addtolength{\tabcolsep}{-3pt} 

\addtolength{\tabcolsep}{+4pt}
\begin{table}[!h]
  \caption{AMIE Dataset Statistics: Slots and Intent Keywords}
  \label{D2}
  \centering
  \begin{tabular}{*4c}
    \toprule
    \textbf{Slot Type} & \textbf{Slot Count} & \textbf{Keyword Type} & \textbf{Keyword Count} \\
    \toprule
    %\multirow{1}[1]{*}{Location} & \multirow{1}[1]{*}{4460} & Intent & 5921 \\
    Location & 4460 & Intent & 5921 \\
    %\multirow{1}[1]{*}{Position/Direction} & \multirow{1}[1]{*}{3187} & Non-Intent & 25000 \\
    Position/Direction & 3187 & Non-Intent & 25000 \\
    %\cmidrule(lr){3-4} 
    \cline{3-4}
    %\multirow{1}[1]{*}{Person} & \multirow{1}[1]{*}{1360} & Valid-Slot & 10954 \\
    Person & 1360 & Valid-Slot & 10954 \\
    %\multirow{1}[1]{*}{Object} & \multirow{1}[1]{*}{632} & Non-Slot & 19967 \\
    Object & 632 & Non-Slot & 19967 \\
    %\cmidrule(lr){3-4}
    \cline{3-4}
    %\multirow{1}[1]{*}{Time Guidance} & \multirow{1}[1]{*}{792} & Intent $\cup$ Valid-Slot & 16875 \\
    Time Guidance & 792 & Intent $\cup$ Valid-Slot & 16875 \\
    %\multirow{1}[2]{*}{Gesture/Gaze} & \multirow{1}[2]{*}{523} & Non-Intent $\cap$ Non-Slot & 14046 \\
    Gesture/Gaze & 523 & Non-Intent $\cap$ Non-Slot & 14046 \\
    %\cline{3-4}
    %\multirow{1}[3]{*}{None} & \multirow{1}[3]{*}{19967} \\
    None & 19967 \\
    %None & 19967 & \textit{Overall} & \textit{30921}\\
    \midrule
    \textit{Total} & \textit{30921} & \textit{Total} & \textit{30921} \\
    \bottomrule
  \end{tabular}
\end{table}
\addtolength{\tabcolsep}{-4pt}

\subsection{Detecting Utterance-level Intent Types}
As a baseline system, we implemented term-frequency and rule-based mapping mechanisms from word-level intent keywords extraction to utterance-level intent recognition. To further improve the utterance-level performance, we explored various RNN architectures and developed a hierarchical (2-level) models to recognize passenger intents along with relevant entities/slots in utterances. Our hierarchical model has the following 2-levels:

\begin{itemize}
%\item (1) Word-level Extraction (to eliminate non-slot \& non-intent keywords first)
%\item (2) Utterance-level Detection (to recognize intent types)
\item Level-1: Word-level extraction (to automatically detect/predict and eliminate non-slot \& non-intent keywords first, as they would not carry much information for understanding the utterance-level intent-type).
\item Level-2: Utterance-level recognition (to detect final intent-types from given utterances, by using valid slots and intent keywords as inputs only, which are detected at Level-1).
\end{itemize}

\subsubsection{RNN with LSTM Cells for Sequence Modeling.}
In this study, we employed an RNN architecture with LSTM cells that are designed to exploit long range dependencies in sequential data. LSTM has memory cell state to store relevant information and various gates, which can mitigate the vanishing gradient problem \cite{lstm-1997}. Given the input $x_t$ at time $t$, and hidden state from the previous time step $h_{t-1}$, the hidden and output layers for the current time step are computed. The LSTM architecture is specified by the following equations:
\begin{align}
%i_t &= \mathrm{sigm} (W_{xi} x_t + W_{hi} h_{t-1} + b_i)\\
%f_t &= \mathrm{sigm} (W_{xf} x_t + W_{hf} h_{t-1} + b_f)\\
%o_t &= \mathrm{sigm} (W_{xo} x_t + W_{ho} h_{t-1} + b_o)\\
i_t &= \sigma (W_{xi} x_t + W_{hi} h_{t-1} + b_i)\\
f_t &= \sigma (W_{xf} x_t + W_{hf} h_{t-1} + b_f)\\
o_t &= \sigma (W_{xo} x_t + W_{ho} h_{t-1} + b_o)\\
g_t &= \tanh (W_{xg} x_t + W_{hg} h_{t-1} + b_g)\\
c_t &= f_t \odot c_{t-1} + i_t \odot g_t \\
h_t &= o_t \odot \tanh (c_t)
\end{align}
where $W$ and $b$ denote the weight matrices and bias terms, respectively. The sigmoid ($\sigma$) and $\tanh$ are activation functions applied element-wise, and $\odot$ denotes the element-wise vector product. LSTM has a memory vector $c_t$ to read/write or reset using a gating mechanism and activation functions. Here, input gate $i_t$ scales down the input, the forget gate $f_t$ scales down the memory vector $c_t$, and the output gate $o_t$ scales down the output to achieve final $h_t$, which is used to predict $y_t$ (through a $softmax$ activation). Similar to LSTMs, GRUs \cite{gru-2014} are proposed as a simpler and faster alternative, having only reset and update gates. For Bi-LSTM \cite{bi-lstm-1997,hakkani-2016}, two LSTM architectures are traversed in forward and backward directions, where their hidden layers are concatenated to compute the output.

\subsubsection{Extracting Slots and Intent Keywords.}
\begin{figure}[t]
\centering
\includegraphics[width=\textwidth]{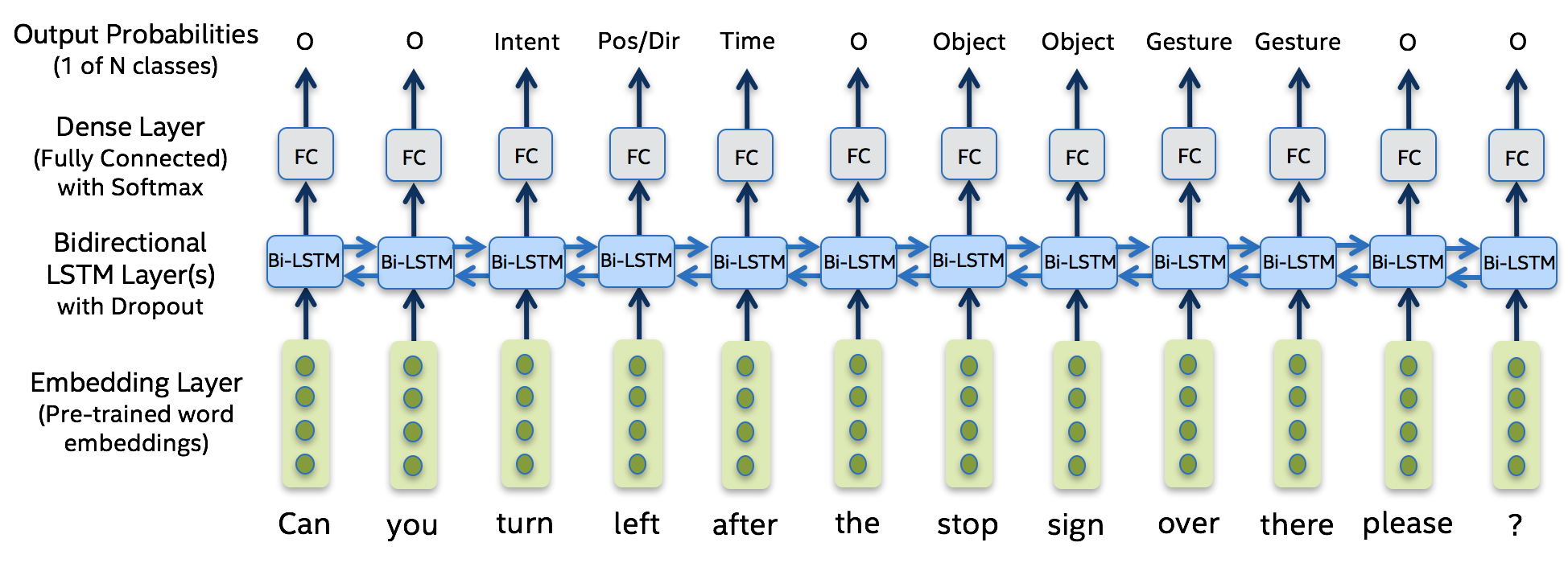}
\caption{Seq2seq Bi-LSTM Network for Slot Filling and Intent Keyword Extraction} \label{fig:network_updated}
\end{figure}

For slot filling and intent keywords extraction, we experimented with various configurations of seq2seq LSTMs \cite{lstm-2015} and GRUs \cite{gru-2014}, as well as Bi-LSTMs \cite{bi-lstm-1997}. A sample network architecture can be seen in Fig.~\ref{fig:network_updated} where we jointly trained slots and intent keywords. The passenger utterance is fed into LSTM/GRU network via an embedding layer as a sequence of words, which are transformed into word vectors. We also experimented with GloVe \cite{glove-2014}, word2vec \cite{word2vec-2013,word2vec-nips-2013}, and fastText \cite{fasttext-2017} as pre-trained word embeddings. To prevent overfitting, we used a dropout layer with~0.5 rate for regularization. Best performing results are obtained with Bi-LSTMs and GloVe embeddings (6B tokens, 400K vocabulary size, vector dimension 100).

\subsubsection{Utterance-level Recognition.}
For utterance-level intent detection, we mainly experimented with 5 groups of models: (1) Hybrid: RNN + Rule-based, (2) Separate: Seq2one Bi-LSTM with Attention, (3) Joint: Seq2seq Bi-LSTM for slots/intent keywords \& utterance-level intents, (4) Hierarchical \& Separate, (5) Hierarchical \& Joint. For (1), we detect/extract intent keywords and slots (via RNN) and map them into utterance-level intent-types (rule-based). For (2), we feed the whole utterance as input sequence and intent-type as single target into Bi-LSTM network with Attention mechanism. For (3), we jointly train word-level intent keywords/slots and utterance-level intents (by adding \textit{\textless BOU\textgreater/\textless EOU\textgreater} terms to the beginning/end of utterances with intent-types as their labels). For (4) and (5), we detect/extract intent keywords/slots first, and then only feed the predicted keywords/slots as a sequence into (2) and (3), respectively. %Further details of these models will be discussed in Section~\ref{exp_res1}.%the Experiments and Results section. %The details of those models are given in the following section.

\section{Experiments and Results}
\label{exp_res}

\subsection{Utterance-Level Intent Detection Experiments}
\label{exp_res1}

The details of 5 groups of models and their variations that we experimented with for utterance-level intent recognition are summarized in this section.

\begin{figure}[t!]
\centering
\includegraphics[scale=0.32]{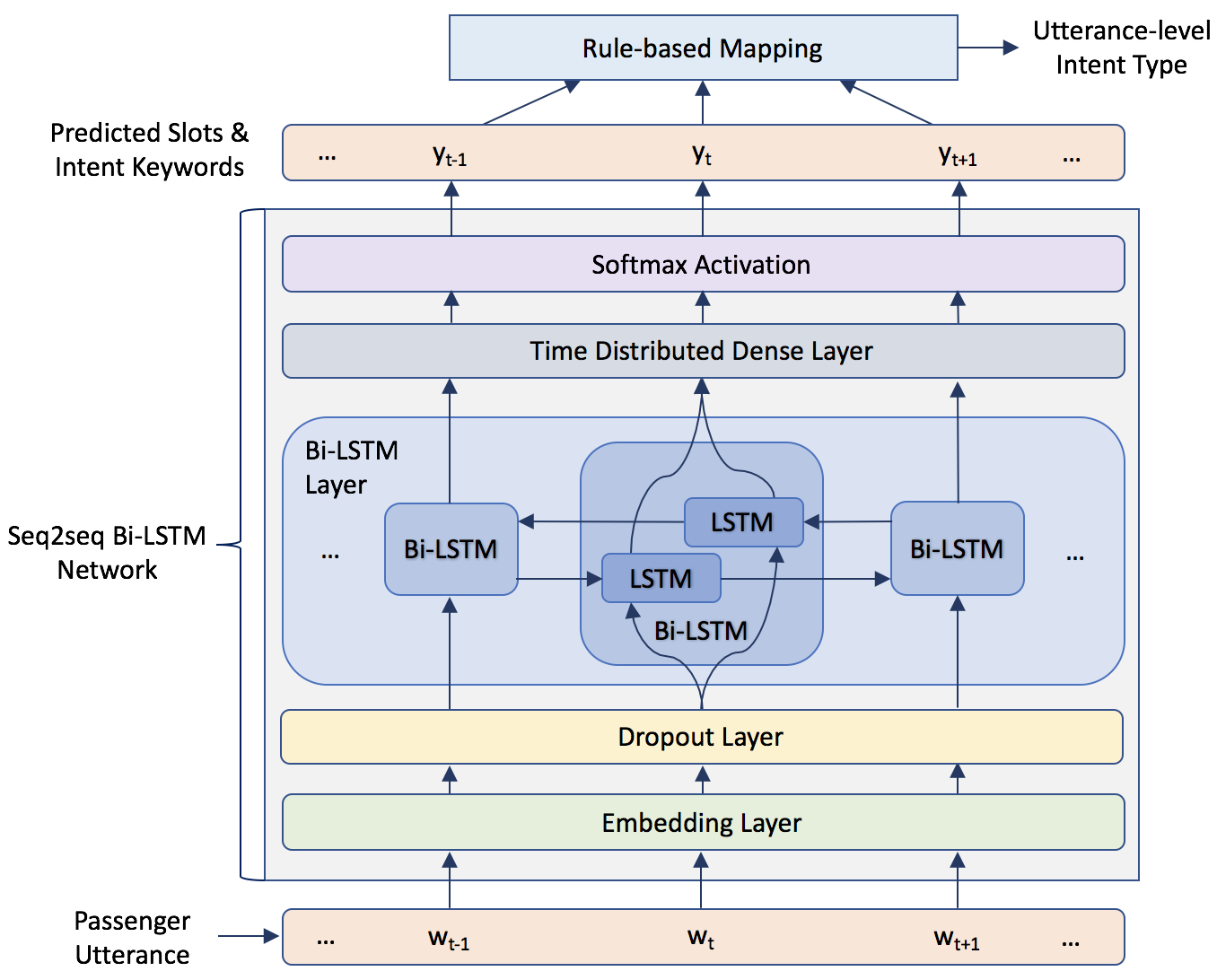}
\caption{Hybrid Models Network Architecture} \label{fig:hybrid}
\end{figure}

\subsubsection{Hybrid Models.} Instead of purely relying on machine learning (ML) or deep learning (DL) system, hybrid models leverage both ML/DL and rule-based systems. In this model, we defined our hybrid approach as using RNNs first for detecting/extracting \textit{intent keywords} and \textit{slots}; then applying rule-based mapping mechanisms to identify \textit{utterance-level intents} (using the predicted \textit{intent keywords} and \textit{slots}). A sample network architecture can be seen in Fig.~\ref{fig:hybrid} where we leveraged seq2seq Bi-LSTM networks for word-level extraction before the rule-based mapping to utterance-level intent classes. The model variations are defined based on varying mapping mechanisms and networks as follows:
\begin{itemize}
\item Hybrid-0: RNN (Seq2seq LSTM for \textit{intent keywords} extraction) + Rule-based (mapping extracted \textit{intent keywords} to \textit{utterance-level intents})
\item Hybrid-1: RNN (Seq2seq Bi-LSTM for \textit{intent keywords} extraction) + Rule-based (mapping extracted \textit{intent keywords} to \textit{utterance-level intents})
\item Hybrid-2: RNN (Seq2seq Bi-LSTM for \textit{intent keywords} \& \textit{slots} extraction) + Rule-based (mapping extracted \textit{intent keywords} \& ‘\textit{Position/Direction}’ \textit{slots} to \textit{utterance-level intents})
\item Hybrid-3: RNN (Seq2seq Bi-LSTM for \textit{intent keywords} \& \textit{slots} extraction) + Rule-based (mapping extracted \textit{intent keywords} \& \textit{all slots} to \textit{utterance-level intents})
\end{itemize}

\begin{figure}[h!]
\centering
%\subfloat[Separate Seq2one Network]{\includegraphics[width=0.57\textwidth]{separate}}
%\subfloat[Separate Seq2one Network]{\includegraphics[width=0.57\textwidth]{separate_updated}}
\subfloat[Separate Seq2one Network]{\includegraphics[width=0.52\textwidth]{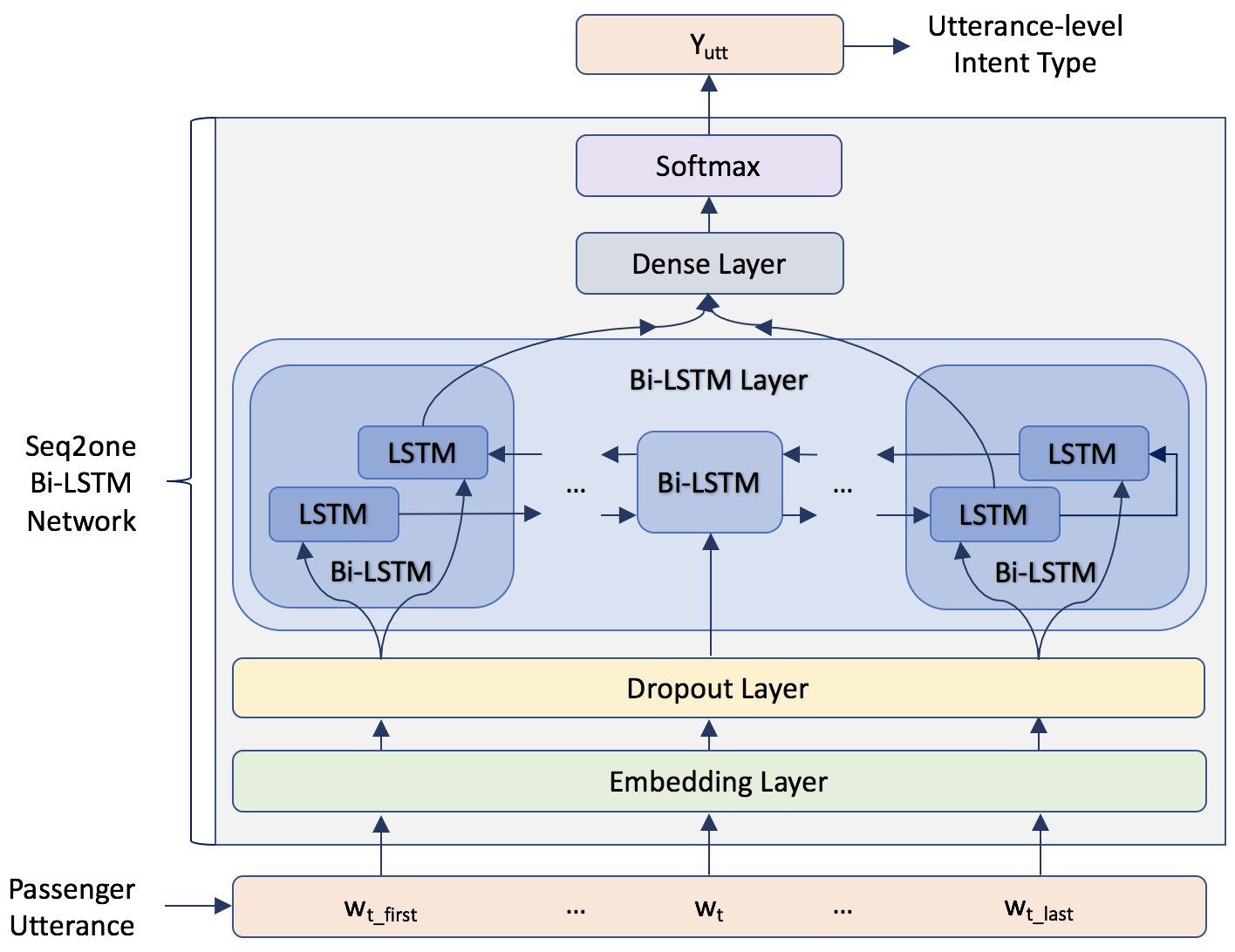}}
\hfill
%\subfloat[Separate Seq2one with Attention]{\includegraphics[width=0.43\textwidth]{separate-att}}
\subfloat[Separate Seq2one with Attention]{\includegraphics[width=0.47\textwidth]{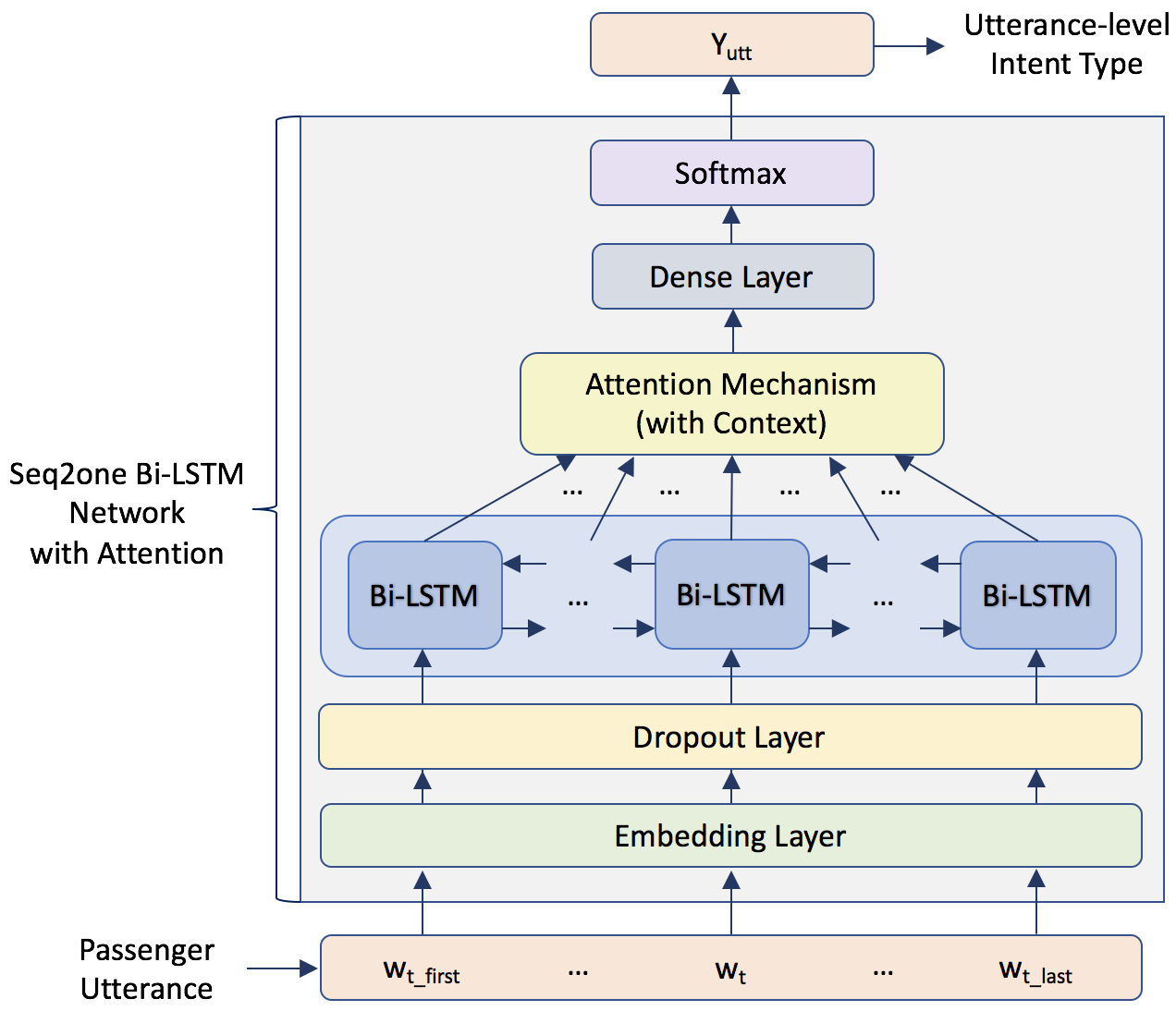}}
\caption{Separate Models Network Architecture}
\label{fig:separate}
\end{figure}

\subsubsection{Separate Seq2one Models.} This approach is based on separately training sequence-to-one RNNs for \textit{utterance-level intents} only. These are called separate models as we do not leverage any information from the \textit{slot} or \textit{intent keyword} tags (i.e., \textit{utterance-level intents} are not jointly trained with \textit{slots}/\textit{intent keywords}). Note that in seq2one models, we feed the utterance as an input sequence and the LSTM layer will only return the hidden state output at the last time step. This single output (or concatenated output of last hidden states from the forward and backward LSTMs in Bi-LSTM case) will be used to classify the intent type of the given utterance. The idea behind is that the last hidden state of the sequence will contain a latent semantic representation of the whole input utterance, which can be utilized for utterance-level intent prediction. See Fig.~\ref{fig:separate} (a) for sample network architecture of the seq2one Bi-LSTM network. Note that in the Bi-LSTM implementation for seq2one learning (i.e., when not returning sequences), the outputs of backward/reverse LSTM is actually ordered in reverse time steps ($t_{last}$ ... $t_{first}$). Thus, as illustrated in Fig.~\ref{fig:separate} (a), we actually concatenate the hidden state outputs of forward LSTM at last time step and backward LSTM at first time step (i.e., first word in a given utterance), and then feed this merged result to the dense layer. Fig.~\ref{fig:separate} (b) depicts the seq2one Bi-LSTM network with Attention mechanism applied on top of Bi-LSTM layers. For the Attention case, the hidden state outputs of all time steps are fed into the Attention mechanism that will allow to point at specific words in a sequence when computing a single output \cite{attention-2015}. Another variation of Attention mechanism we examined is the AttentionWithContext, which incorporates a context/query vector jointly learned during the training process to assist the attention \cite{attention-2016}. All seq2one model variations we experimented with can be summarized as follows:
\begin{itemize}
\item Separate-0: Seq2one LSTM for \textit{utterance-level intents}
\item Separate-1: Seq2one Bi-LSTM for \textit{utterance-level intents}
\item Separate-2: Seq2one Bi-LSTM with Attention \cite{attention-2015} for \textit{utterance-level intents}
\item Separate-3: Seq2one Bi-LSTM with AttentionWithContext \cite{attention-2016} for \textit{utterance-level intents} 
\end{itemize}

\begin{figure}[h!]
\centering
\includegraphics[scale=0.32]{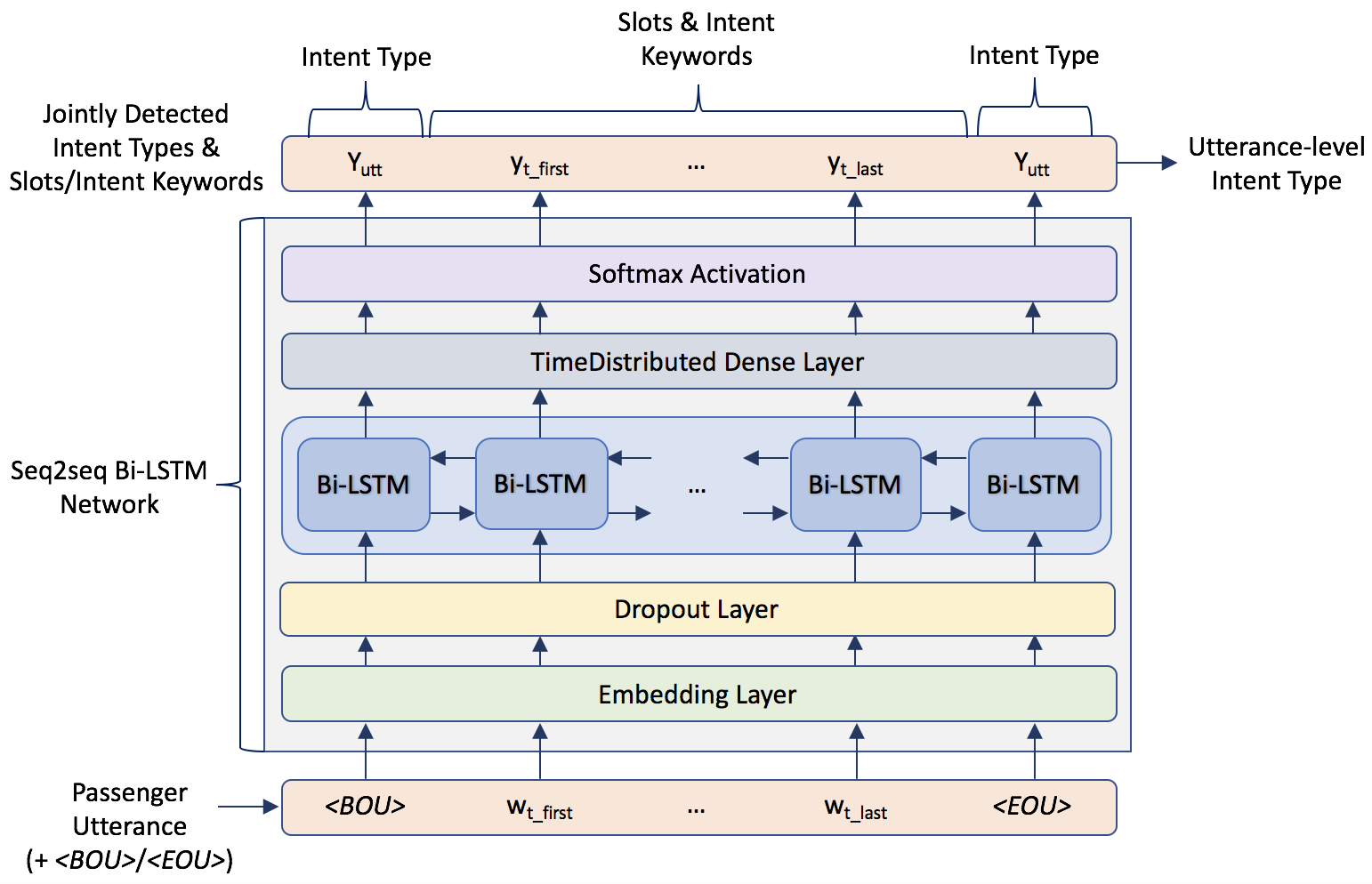}
\caption{Joint Models Network Architecture} \label{fig:joint}
\end{figure}

\subsubsection{Joint Seq2seq Models.} Using sequence-to-sequence networks, the approach here is jointly training annotated \textit{utterance-level intents} and \textit{slots}/\textit{intent keywords} by adding \textit{\textless BOU\textgreater/ \textless EOU\textgreater} tokens to the beginning/end of each utterance, with \textit{utterance-level intent-type} as labels of such tokens. Our approach is an extension of \cite{hakkani-2016}, in which only an \textit{\textless EOS\textgreater} term is added with \textit{intent-type} tags associated to this sentence final token, both for LSTM and Bi-LSTM cases. However, we experimented with adding both \textit{\textless BOU\textgreater} and \textit{\textless EOU\textgreater} terms as Bi-LSTMs will be used for seq2seq learning, and we observed that slightly better results can be achieved by doing so. The idea behind is that, since this is a seq2seq learning problem, at the last time step (i.e., prediction at \textit{\textless EOU\textgreater}) the reverse pass in Bi-LSTM would be incomplete (refer to Fig.~\ref{fig:separate} (a) to observe the last Bi-LSTM cell). Therefore, adding \textit{\textless BOU\textgreater} token and leveraging the backward LSTM output at first time step (i.e., prediction at \textit{\textless BOU\textgreater}) would potentially help for joint seq2seq learning. An overall network architecture can be found in Fig.~\ref{fig:joint} for our joint models. We will report the experimental results on two variations (with and without \textit{intent keywords}) as follows:
\begin{itemize}
\item Joint-1: Seq2seq Bi-LSTM for \textit{utterance-level intent} detection (jointly trained with \textit{slots})
\item Joint-2: Seq2seq Bi-LSTM for \textit{utterance-level intent} detection (jointly trained with \textit{slots} \& \textit{intent keywords}) 
\end{itemize}

\begin{figure}[h!]
\centering
\subfloat[Hierarchical \& Separate Model]{\includegraphics[width=0.48\textwidth]{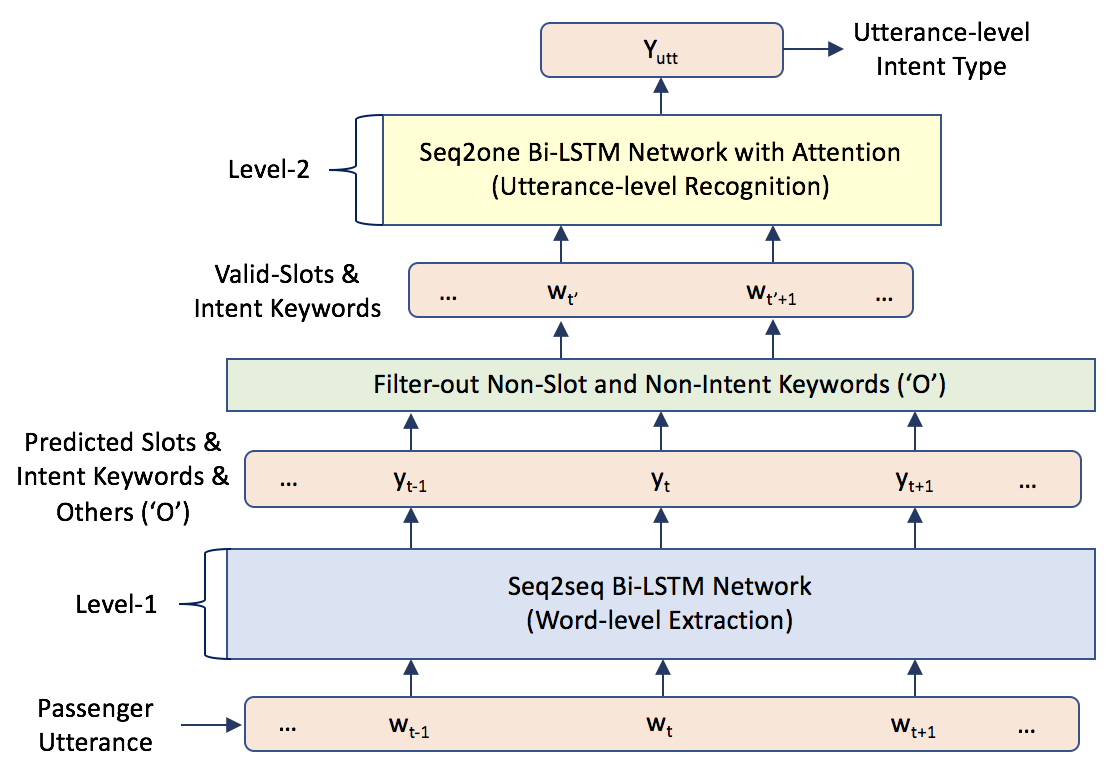}}
%\quad
\hfill
\subfloat[Hierarchical \& Joint Model]{\includegraphics[width=0.51\textwidth]{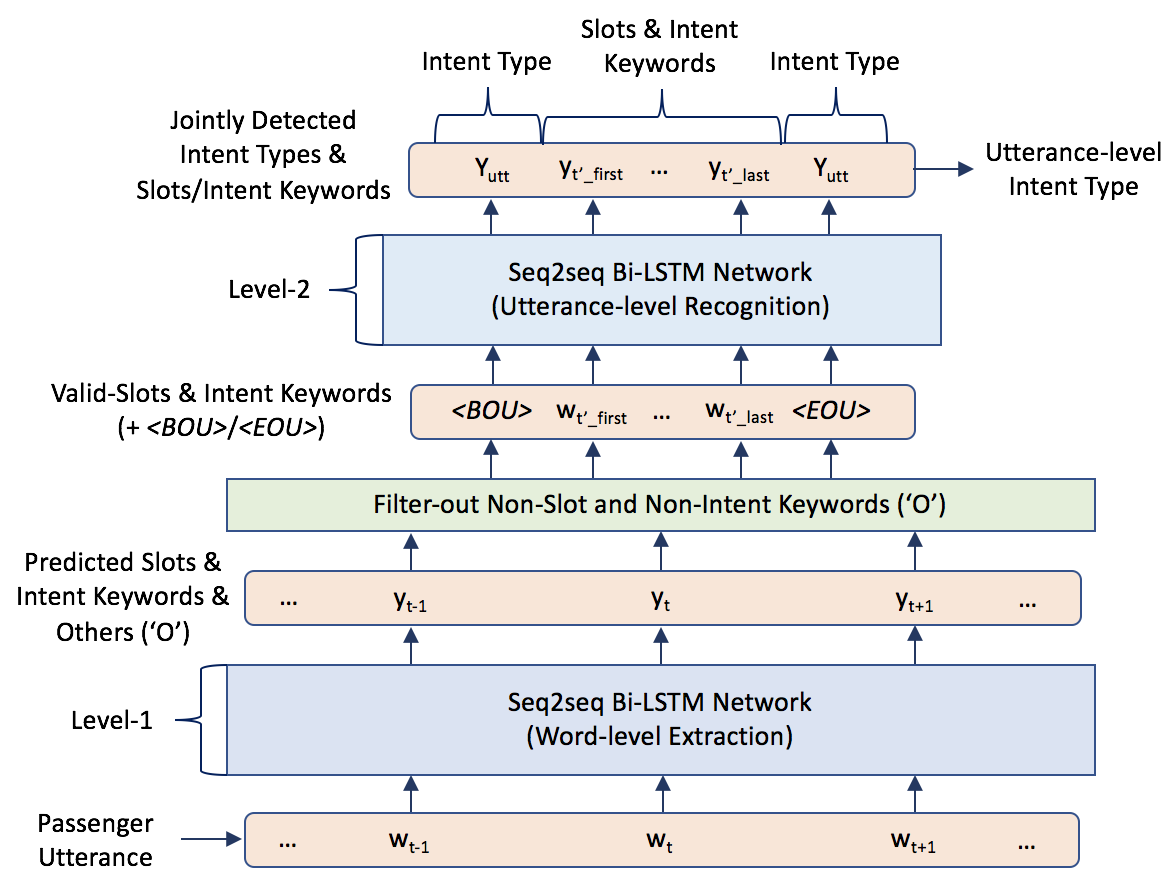}}
\caption{Hierarchical Models Network Architecture}
\label{fig:hierarchical}
\end{figure}

\subsubsection{Hierarchical \& Separate Models.} Proposed hierarchical models are detecting/extracting \textit{intent keywords} \& \textit{slots} using sequence-to-sequence networks first (i.e., level-1), and then feeding only the words that are predicted as \textit{intent keywords} \& \textit{valid slots} (i.e., not the ones that are predicted as ‘\textit{None/O}’) as an input sequence to various separate sequence-to-one models (described above) to recognize final \textit{utterance-level intents} (i.e., level-2). A sample network architecture is given in Fig.~\ref{fig:hierarchical} (a). The idea behind filtering out non-slot and non-intent keywords here resembles providing a summary of input sequence to the upper levels of the network hierarchy, where we actually learn this summarized sequence of keywords using another RNN layer. This would potentially result in focusing the utterance-level classification problem on the most salient words of the input sequences (i.e., \textit{intent keywords} \& \textit{slots}) and also effectively reducing the length of input sequences (i.e., improving the long-term dependency issues observed in longer sequences). Note that according to our dataset statistics given in Table~\ref{D2},~45\% of the words found in transcribed utterances with passenger intents are annotated as non-slot and non-intent keywords (e.g., 'please', 'okay', 'can', 'could', incomplete/interrupted words, filler sounds like 'uh'/'um', certain stop words, punctuation, and many others that are not related to intent/slots). Therefore, the proposed approach would result in reducing the sequence length nearly by half at the input layer of level-2 for utterance-level recognition. For hierarchical \& separate models, we experimented with 4 variations based on which separate model used at the second level of the hierarchy, and these are summarized as follows:
\begin{itemize}
%\item Level-1: Seq2seq Bi-LSTM (for \textit{intent keywords} \& \textit{slots} extraction) + \\Level-2: Separate-0/1/2/3 Seq2one Models (for \textit{utterance-level intents})
%\item Hierarchical \& Separate-0/1/2/3: \\Level-1 (Seq2seq Bi-LSTM for \textit{intent keywords} \& \textit{slots} extraction) + \\Level-2 (Separate-0/1/2/3: Seq2one models for \textit{utterance-level intents})
\item Hierarchical \& Separate-0: Level-1 (Seq2seq LSTM for \textit{intent keywords} \& \textit{slots} extraction) + Level-2 (Separate-0: Seq2one LSTM for \textit{utterance-level intent} detection)
\item Hierarchical \& Separate-1: Level-1 (Seq2seq Bi-LSTM for \textit{intent keywords} \& \textit{slots} extraction) + Level-2 (Separate-1: Seq2one Bi-LSTM for \textit{utterance-level intent} detection)
\item Hierarchical \& Separate-2: Level-1 (Seq2seq Bi-LSTM for \textit{intent keywords} \& \textit{slots} extraction) + Level-2 (Separate-2: Seq2one Bi-LSTM + Attention for \textit{utterance-level intent} detection)
\item Hierarchical \& Separate-3: Level-1 (Seq2seq Bi-LSTM for \textit{intent keywords}~\&
\textit{slots} extraction) + Level-2 (Separate-3: Seq2one Bi-LSTM + AttentionWithContext for \textit{utterance-level intent} detection)
\end{itemize}

\subsubsection{Hierarchical \& Joint Models.} Proposed hierarchical models detect/extract \textit{intent keywords} \& \textit{slots} using sequence-to-sequence networks first, and then only the words that are predicted as \textit{intent keywords} \& \textit{valid slots} (i.e., not the ones that are predicted as ‘\textit{None/O}’) are fed as input to the joint sequence-to-sequence models (described above). See Fig.~\ref{fig:hierarchical} (b) for sample network architecture. After the filtering or summarization of sequence at level-1, \textit{\textless BOU\textgreater} and \textit{\textless EOU\textgreater} tokens are appended to the shorter input sequence before level-2 for joint learning. Note that in this case, using Joint-1 model (jointly training annotated \textit{slots} \& \textit{utterance-level intents}) for the second level of the hierarchy would not make much sense (without \textit{intent keywords}). Hence, Joint-2 model is used for the second level as described below:
\begin{itemize}
%\item Level-1 (Seq2seq Bi-LSTM for \textit{intent keywords} \& \textit{slots} extraction) + \\Level-2 (Joint-2 Seq2seq models with \textit{slots} \& \textit{intent keywords} \& \textit{utterance-level intents})
\item Hierarchical \& Joint-2: Level-1 (Seq2seq Bi-LSTM for \textit{intent keywords} \& \textit{slots} extraction) + Level-2 (Joint-2 Seq2seq models with \textit{slots} \& \textit{intent keywords} \& \textit{utterance-level intents})
\end{itemize}

%Table~\ref{T1} summarizes the results of various approaches we investigated for utterance-level intent understanding. We achieved 0.91 overall F1-score with our best-performing models (Hierarchical \& Joint-2). We compared our results with the Dialogflow using recommended hybrid mode (rule-based and ML). The same AMIE dataset and annotations are used to train and test (via 10-fold cross-validation) Dialogflow's intent recognition and entity/slot filling modules. Note that with Dialogflow's Detect Intent API, we could reach 0.89 overall F1-score for the same task. %Table~\ref{T2} shows the intent-wise detection results for the AMIE scenarios with our initial (Hybrid-0) and currently best performing (H-Joint-2) intent recognizers. As you can see, our hierarchical \& joint models obtained higher results than the Dialogflow for 8 out of 10 types of intents. More importantly, previously problematic \textit{Destination} and \textit{Route} intents’ detection performances in baseline models (Hybrid-0) jumped from 0.78 to 0.89 and 0.75 to 0.88, respectively.

%Table~\ref{T2} shows the intent-wise detection results for the AMIE scenarios with our initial (Hybrid-0) and currently best performing (H-Joint-2) intent recognizers. As you can see, our hierarchical \& joint models obtained higher results than the Dialogflow for 8 out of 10 types of intents. More importantly, previously problematic \textit{Destination} and \textit{Route} intents’ detection performances in baseline models (Hybrid-0) jumped from 0.78 to 0.89 and 0.75 to 0.88, respectively.

Table~\ref{T1} summarizes the results of various approaches we investigated for utterance-level intent understanding. We achieved 0.91 overall F1-score with our best-performing model, namely \textit{Hierarchical \& Joint-2}. All model results are obtained via 10-fold cross-validation (10-fold CV) on the same dataset. For our AMIE scenarios, Table~\ref{T2} shows the intent-wise detection results with the initial (\textit{Hybrid-0}) and currently best performing (\textit{H-Joint-2}) intent recognizers. With our best model (\textit{H-Joint-2}), relatively problematic \textit{SetDestination} and \textit{SetRoute} intents’ detection performances in baseline model (\textit{Hybrid-0}) jumped from 0.78 to 0.89 and 0.75 to 0.88, respectively.

We compared our intent detection results with the Dialogflow's Detect Intent API. The same AMIE dataset is used to train and test (10-fold CV) Dialogflow's intent detection and slot filling modules, using the recommended hybrid mode (rule-based and ML). As shown in Table~\ref{T2}, an overall F1-score of 0.89 is achieved with Dialogflow for the same task. As you can see, our \textit{Hierarchical \& Joint} models obtained higher results than the Dialogflow for 8 out of 10 intent types. 

\addtolength{\tabcolsep}{+3pt} 
\begin{table}[b!]
  \caption{Utterance-level Intent Detection Performance Results (10-fold CV)}
  \label{T1}
  \centering
  \resizebox{\columnwidth}{!}{
  \begin{tabular}{lccc}
    \toprule
    %\multicolumn{1}{c}{Model Type}                   \\
    %\cmidrule(r){1-2}
    \textbf{Model Type} & \textbf{\textit{Prec}} & \textbf{\textit{Rec}} & \textbf{F1} \\
    %\midrule
    \toprule
    Hybrid-0: RNN (LSTM) + Rule-based (\textit{intent keywords}) & \textit{0.86} & \textit{0.85} & 0.85 \\
    Hybrid-1: RNN (Bi-LSTM) + Rule-based (\textit{intent keywords}) & \textit{0.87} & \textit{0.86} & 0.86 \\
    Hybrid-2: RNN (Bi-LSTM) + Rule-based (\textit{intent keywords \& Pos slots}) & \textit{0.89} & \textit{0.88} & 0.88 \\
    Hybrid-3: RNN (Bi-LSTM) + Rule-based (\textit{intent keywords \& all slots}) & \textit{0.90} & \textit{0.90} & 0.90 \\
    \midrule
    Separate-0: Seq2one LSTM & \textit{0.87} & \textit{0.86} & 0.86 \\
    Separate-1: Seq2one Bi-LSTM & \textit{0.88} & \textit{0.88} & 0.88 \\
    Separate-2: Seq2one Bi-LSTM + Attention & \textit{0.88} & \textit{0.88} & 0.88 \\
    Separate-3: Seq2one Bi-LSTM + AttentionWithContext & \textit{0.89} & \textit{0.89} & 0.89 \\
    \midrule
    Joint-1: Seq2seq Bi-LSTM (\textit{uttr-level intents \&  slots}) & \textit{0.88} & \textit{0.87} & 0.87 \\
    Joint-2: Seq2seq Bi-LSTM (\textit{uttr-level intents \&  slots \& intent keywords}) & \textit{0.89} & \textit{0.88} & 0.88 \\
    \midrule
    Hierarchical \& Separate-0 (LSTM) & \textit{0.88} & \textit{0.87} & 0.87 \\
    Hierarchical \& Separate-1 (Bi-LSTM) & \textit{0.90} & \textit{0.90} & 0.90 \\
    Hierarchical \& Separate-2 (Bi-LSTM + Attention) & \textit{0.90} & \textit{0.90} & 0.90 \\
    Hierarchical \& Separate-3 (Bi-LSTM + AttentionWithContext) & \textit{0.90} & \textit{0.90} & 0.90 \\
    \midrule
    Hierarchical \& Joint-2 (\textit{uttr-level intents \&  slots \& intent keywords}) & \textit{0.91} & \textit{0.90} & \textbf{0.91} \\
    \bottomrule
  \end{tabular}
  }
\end{table}
\addtolength{\tabcolsep}{-3pt} 

\iffalse
%\begin{table*}[t]
\begin{table}[b!]
  \caption{Utterance-level Intent Detection Performance Results (10-fold CV)}
  \label{T1}
  \centering
  \resizebox{\columnwidth}{!}{
  \begin{tabular}{lc}
    \toprule
    %\multicolumn{1}{c}{Model Type}                   \\
    %\cmidrule(r){1-2}
    \textbf{Model Type} & \textbf{F1} \\
    %\midrule
    \toprule
    Hybrid-0: RNN (LSTM) + Rule-based (via \textit{intent keywords}) & 0.85 \\
    Hybrid-1: RNN (Bi-LSTM) + Rule-based (via \textit{intent keywords}) & 0.86 \\
    Hybrid-2: RNN (Bi-LSTM) + Rule-based (via \textit{intent keywords \& Pos/Dir slots}) & 0.88 \\
    Hybrid-3: RNN (Bi-LSTM) + Rule-based (via \textit{intent keywords \& all slots}) & 0.90\\
    \midrule
    Separate-0: Seq2one LSTM & 0.86 \\
    Separate-1: Seq2one Bi-LSTM & 0.88 \\
    Separate-2: Seq2one Bi-LSTM + Attention & 0.88 \\
    Separate-3: Seq2one Bi-LSTM + AttentionWithContext & 0.89\\
    \midrule
    Joint-1: Seq2seq Bi-LSTM (\textit{utterance-level intent-types \&  slots}) & 0.87 \\
    Joint-2: Seq2seq Bi-LSTM (\textit{utterance-level intent-types \&  slots \& intent keywords}) & 0.88 \\
    \midrule
    Hierarchical \& Separate-0 (LSTM) & 0.87 \\
    Hierarchical \& Separate-1 (Bi-LSTM) & 0.90 \\
    Hierarchical \& Separate-2 (Bi-LSTM + Attention) & 0.90 \\
    Hierarchical \& Separate-3 (Bi-LSTM + AttentionWithContext) & 0.90\\
    \midrule
    Hierarchical \& Joint-2 (\textit{utterance-level intent-types \&  slots \& intent keywords}) & \textbf{0.91} \\
    \bottomrule
  \end{tabular}
  }
\end{table}
\fi

\addtolength{\tabcolsep}{+3pt}   
\begin{table}[t!]
  \caption{Intent-wise Performance Results of Utterance-level Intent Detection} %(10-fold CV)}
  \label{T2}
  \centering
  \resizebox{\columnwidth}{!}{
  \begin{tabular}{*{11}c}
    \toprule
     \textbf{AMIE} & \textbf{Intent} & \multicolumn{6}{c}{\textbf{Our Intent Detection Models}} & \multicolumn{3}{c}{\textbf{Dialogflow}} \\
     \textbf{Scenario} & \textbf{Type} & \multicolumn{3}{c}{\textbf{Baseline} (Hybrid-0)} & \multicolumn{3}{c}{\textbf{Best} (H-Joint-2)} & \multicolumn{3}{c}{\textbf{Intent Detection}} \\
     % & & \multicolumn{6}{c}{\textbf{Our Intent Detection Models}} & \multicolumn{3}{c}{\textbf{Dialogflow}} \\
     %\textbf{AMIE} & \textbf{Intent} & \multicolumn{3}{c}{\textbf{Baseline}} & \multicolumn{3}{c}{\textbf{Best}} & \multicolumn{3}{c}{\textbf{Intent}} \\
     %\textbf{Scenario} & \textbf{Type} & \multicolumn{3}{c}{(Hybrid-0)} & \multicolumn{3}{c}{(H-Joint-2)} & \multicolumn{3}{c}{\textbf{Detection}} \\
    %\midrule
    \cmidrule(lr){3-5} \cmidrule(lr){6-8} \cmidrule(lr){9-11}
      & & \textit{Prec} & \textit{Rec} & \textbf{F1} & \textit{Prec} & \textit{Rec} & \textbf{F1} & \textit{Prec} & \textit{Rec} & \textbf{F1} \\
    \toprule
    Finishing & Stop & \textit{0.88} & \textit{0.91} & 0.90 & \textit{0.93} & \textit{0.91} & 0.92 & \textit{0.89} & \textit{0.90} & 0.90 \\
    the Trip & Park & \textit{0.96} & \textit{0.87} & 0.91 & \textit{0.94} & \textit{0.94} & 0.94 & \textit{0.95} & \textit{0.88} & 0.91 \\
     & PullOver & \textit{0.95} & \textit{0.96} & 0.95 & \textit{0.97} & \textit{0.94} & 0.96 & \textit{0.95} & \textit{0.97} & 0.96 \\
     & DropOff & \textit{0.90} & \textit{0.95} & 0.92 & \textit{0.95} & \textit{0.95} & 0.95 & \textit{0.96} & \textit{0.91} & 0.93 \\
    \midrule
    Dest/Route & SetDest & \textit{0.70} & \textit{0.88} & 0.78 & \textit{0.89} & \textit{0.90} & 0.89 & \textit{0.84} & \textit{0.91} & 0.87 \\
     & SetRoute & \textit{0.80} & \textit{0.71} & 0.75 & \textit{0.86} & \textit{0.89} & 0.88 & \textit{0.83} & \textit{0.86} & 0.84 \\
    \midrule
    Speed & GoFaster & \textit{0.86} & \textit{0.89} & 0.88 & \textit{0.89} & \textit{0.90} & 0.90 & \textit{0.94} & \textit{0.92} & 0.93 \\
     & GoSlower & \textit{0.92} & \textit{0.84} & 0.88 & \textit{0.89} & \textit{0.86} & 0.88 & \textit{0.93} & \textit{0.87} & 0.90 \\
    \midrule
    Others & OpenDoor & \textit{0.95} & \textit{0.95} & 0.95 & \textit{0.95} & \textit{0.95} & 0.95 & \textit{0.94} & \textit{0.93} & 0.93 \\
     & Other & \textit{0.92} & \textit{0.72} & 0.80 & \textit{0.83} & \textit{0.81} & 0.82 & \textit{0.88} & \textit{0.73} & 0.80 \\
    \midrule
     & Overall & \textit{0.86} & \textit{0.85} & \textbf{0.85} & \textit{0.91} & \textit{0.90} & \textbf{0.91} & \textit{0.90} & \textit{0.89} & \textbf{0.89} \\
    \bottomrule
  \end{tabular}
  }
\end{table}
\addtolength{\tabcolsep}{-3pt}

\iffalse
\addtolength{\tabcolsep}{+3pt}   
\begin{table}[!t]
  \caption{Intent-wise F1-scores of Utterance-level Intent Detection Models} %(10-fold CV)}
  \label{T2}
  \centering
  %\resizebox{\columnwidth}{!}{
  \begin{tabular}{*5c}
    \toprule
    \textbf{AMIE} & \textbf{Intent} & \multicolumn{2}{c}{\textbf{Our Intent Detection}} & \textbf{Dialogflow} \\
    \textbf{Scenario} & \textbf{Type} & {Baseline} & {Best} & \textbf{Intent} \\
     & & {(Hybrid-0)} & {(H-Joint-2)} & \textbf{Detection} \\
    \toprule
    Finishing-Trip & Stop & 0.90 & 0.92 & 0.90 \\
     & Park & 0.91 & 0.94 & 0.91 \\
     & PullOver & 0.95 & 0.96 & 0.96 \\
     & DropOff & 0.92 & 0.95 & 0.93 \\
    \midrule
    Destination/Route & SetDest & 0.78 & 0.89 & 0.87 \\
     & SetRoute & 0.75 & 0.88 & 0.84 \\
    \midrule
    Behavior/Speed & GoFaster & 0.88 & 0.90 & 0.93 \\
     & GoSlower & 0.88 & 0.88 & 0.90 \\
    \midrule
    Others & OpenDoor & 0.95 & 0.95 & 0.93 \\
     & Other & 0.80 & 0.82 & 0.80 \\
    \midrule
     & \textit{Overall} & \textit{0.85} & \textit{\textbf{0.91}} & \textbf{\textit{0.89}} \\
    \bottomrule
  \end{tabular}
  %}
\end{table}
\addtolength{\tabcolsep}{-3pt}
\fi

\addtolength{\tabcolsep}{+3pt}
\begin{table}[b!]
  \caption{Slot Filling Results (10-fold CV)}
  \label{T3}
  \centering
  \resizebox{\columnwidth}{!}{
  \begin{tabular}{*{10}c}
    \toprule
    & \multicolumn{6}{c}{\textbf{Our Slot Filling Models}} & \multicolumn{3}{c}{\textbf{Dialogflow}} \\
     & \multicolumn{3}{c}{\textbf{Se2qseq LSTM}} & \multicolumn{3}{c}{\textbf{Se2qseq Bi-LSTM}} & \multicolumn{3}{c}{\textbf{Slot Filling}} \\
    %\midrule
    \cmidrule(lr){2-4} \cmidrule(lr){5-7} \cmidrule(lr){8-10}
     \textbf{Slot Type}  & \textit{Prec} & \textit{Rec} & \textbf{F1} & \textit{Prec} & \textit{Rec} & \textbf{F1} & \textit{Prec} & \textit{Rec} & \textbf{F1}\\
    \toprule
    Location & \textit{0.94} & \textit{0.92} & 0.93 & \textit{0.96} & \textit{0.94} & 0.95 & \textit{0.94} & \textit{0.81} & 0.87 \\
    Position/Direction & \textit{0.92} & \textit{0.93} & 0.93 & \textit{0.95} & \textit{0.95} & 0.95 & \textit{0.91} & \textit{0.92} & 0.91 \\
    Person & \textit{0.97} & \textit{0.96} & 0.97 & \textit{0.98} & \textit{0.97} & 0.97 & \textit{0.96} & \textit{0.76} & 0.85 \\
    Object & \textit{0.82} & \textit{0.79} & 0.80 & \textit{0.93} & \textit{0.85} & 0.89 & \textit{0.96} & \textit{0.70} & 0.81 \\
    Time Guidance & \textit{0.88} & \textit{0.73} & 0.80 & \textit{0.90} & \textit{0.80} & 0.85 & \textit{0.93} & \textit{0.82} & 0.87 \\
    Gesture/Gaze & \textit{0.86} & \textit{0.88} & 0.87 & \textit{0.92} & \textit{0.92} & 0.92 & \textit{0.86} & \textit{0.65} & 0.74 \\
    None & \textit{0.97} & \textit{0.98} & 0.97 & \textit{0.97} & \textit{0.98} & 0.98 & \textit{0.92} & \textit{0.98} & 0.95 \\
    \midrule
    Overall & \textit{0.95} & \textit{0.95} & \textbf{0.95} & \textit{0.96} & \textit{0.96} & \textbf{0.96} & \textit{0.92} & \textit{0.92} & \textbf{0.92} \\
    \bottomrule
  \end{tabular}
  }
\end{table}
\addtolength{\tabcolsep}{-3pt}

\subsection{Slot Filling and Intent Keyword Extraction Experiments}

 Slot filling and intent keyword extraction results are given in Table~\ref{T3} and Table~\ref{T4}, respectively. For slot extraction, we reached 0.96 overall F1-score using seq2seq Bi-LSTM model, which is slightly better than using LSTM model. Although the overall performance is slightly improved with Bi-LSTM model, relatively problematic \textit{Object}, \textit{Time Guidance}, \textit{Gesture/Gaze} slots’ F1-score performances increased from 0.80 to 0.89, 0.80 to 0.85, and 0.87 to 0.92, respectively. Note that with Dialogflow, we reached 0.92 overall F1-score for the entity/slot filling task on the same dataset. As you can see, our models reached significantly higher F1-scores than the Dialogflow for 6 out of 7 slot types (except \textit{Time Guidance}).

\addtolength{\tabcolsep}{+3pt}
\begin{table}[t!]
  \caption{Intent Keyword Extraction Results (10-fold CV)}
  \label{T4}
  \centering
  \begin{tabular}{*4c}
    \toprule
    \textbf{Keyword Type} & \textbf{\textit{Prec}} & \textbf{\textit{Rec}} & \textbf{F1}\\
    \toprule
    Intent & \textit{0.95} & \textit{0.93} & 0.94 \\
    Non-Intent & \textit{0.98} & \textit{0.99} & 0.99 \\
    \midrule
    Overall & \textit{0.98} & \textit{0.98} & \textbf{0.98} \\
    \bottomrule
  \end{tabular}
\end{table}
\addtolength{\tabcolsep}{-3pt}

\subsection{Speech-to-Text Experiments for AMIE: Training and Testing Models on ASR Outputs}

For transcriptions, utterance-level audio clips were extracted from the passenger-facing video stream, which was the single source used for human transcriptions of all utterances from passengers, AMIE agent and the game master. Since our transcriptions-based intent/slot models assumed perfect (at least close to human-level) ASR in the previous sections, we experimented with more realistic scenario of using ASR outputs for intent/slot modeling. We employed Cloud Speech-to-Text API to obtain ASR outputs on audio clips with passenger utterances, which were segmented using transcription time-stamps. We observed an overall word error rate (WER) of 13.6\% in ASR outputs for all 20 sessions of AMIE.
 
Considering that a generic ASR is used with no domain-specific acoustic models for this moving vehicle environment with in-cabin noise, the initial results were quite promising for us to move on with the model training on ASR outputs. For initial explorations, we created a new dataset having utterances with commands using ASR outputs of the in-cabin data (20 sessions with 1331 spoken utterances). Human transcriptions version of this set is also created. Although the dataset size is limited, both slot/intent keyword extraction models and utterance-level intent recognition models are not severely affected when trained and tested (10-fold CV) on ASR outputs instead of manual transcriptions. See Table~\ref{T5} for the overall F1-scores of the compared models.

%\begin{table*}[t]
\begin{table}[h!]
  %\caption{F1-scores of Models Trained \& Tested (10-fold CV) on Transcriptions vs. ASR Outputs}
  \caption{F1-scores of Models Trained/Tested on Transcriptions vs. ASR Outputs}
  \label{T5}
  \centering
  \resizebox{\columnwidth}{!}{
  \begin{tabular}{lcccccc}
    \toprule
    %& \multicolumn{3}{c}{\textbf{Train on Transcriptions}} & \multicolumn{3}{c}{\textbf{Train on ASR Outputs}} \\
    & \multicolumn{3}{c}{\textbf{Train/Test on}} & \multicolumn{3}{c}{\textbf{Train/Test on}} \\
    & \multicolumn{3}{c}{\textbf{Transcriptions}} & \multicolumn{3}{c}{\textbf{ASR Outputs}} \\
    \toprule
    \textbf{Slot Filling \& Intent Keywords} & \textbf{ALL} & \textit{Singleton} & \textit{Dyad} & \textbf{ALL} & \textit{Singleton} & \textit{Dyad} \\
    \midrule
    Slot Filling & 0.97 & \textit{0.96} & \textit{0.96} & 0.95 & \textit{0.94} & \textit{0.93} \\
    Intent Keyword Extraction & 0.98 & \textit{0.98} & \textit{0.97} & 0.97 & \textit{0.96} & \textit{0.96} \\
    Slot Filling \& Intent Keyword Extraction & \textbf{0.95} & \textit{0.95} & \textit{0.94} & \textbf{0.94} & \textit{0.92} & \textit{0.91} \\
    \midrule
    \textbf{Utterance-level Intent Detection} & \textbf{ALL} & \textit{Singleton} & \textit{Dyad} & \textbf{ALL} & \textit{Singleton} & \textit{Dyad} \\
    \midrule
    %Separate & 0.84 & - & - & 0.83 & - & - \\
    %Separate + Attention & 0.87 & - & - & 0.85 & - & - \\
    %\midrule
    %Joint & 0.86 & - & - & 0.84 & - & - \\
    %\midrule
    Hierarchical \& Separate & 0.87 & \textit{0.85} & \textit{0.86} & 0.85 & \textit{0.84} & \textit{0.83}  \\
    Hierarchical \& Separate + Attention & 0.89 & \textit{0.86} & \textit{0.87} & 0.86 & \textit{0.84} & \textit{0.84} \\
    %\midrule
    Hierarchical \& Joint & \textbf{0.89} & \textit{0.87} & \textit{0.88} & \textbf{0.87} & \textit{0.85} & \textit{0.85} \\
    \bottomrule
  \end{tabular}
  }
\end{table}

\subsubsection{Singleton versus Dyad Sessions.}
After the ASR pipeline described above is completed for all 20 sessions of AMIE in-cabin dataset (\textit{ALL} with 1331 utterances), we repeated all our experiments with the subsets for 10 sessions having single passenger (\textit{Singletons} with 600 utterances) and remaining 10 sessions having two passengers (\textit{Dyads} with 731 utterances). We observed overall WER of 13.5\% and 13.7\% for \textit{Singletons} and \textit{Dyads}, respectively. The overlapping speech cases with slightly more conversations going on (longer transcriptions) in \textit{Dyad} sessions compared to the \textit{Singleton} sessions may affect the ASR performance, which may also affect the intent/slots models performances. As shown in Table~\ref{T5}, although we have more samples with \textit{Dyads}, the performance drops between the models trained on transcriptions vs. ASR outputs are slightly higher for the \textit{Dyads} compared to the \textit{Singletons}, as expected.

\section{Discussion and Conclusion}

We introduced AMIE, the intelligent in-cabin car agent responsible for handling certain AV-passenger interactions. We develop hierarchical and joint models to extract various passenger intents along with relevant slots for actions to be performed in AV, achieving F1-scores of 0.91 for intent recognition and 0.96 for slot extraction. We show that even using the generic ASR with noisy outputs, our models are still capable of achieving comparable results with models trained on human transcriptions. We believe that the ASR can be improved by collecting more in-domain data to obtain domain-specific acoustic models. These initial models will allow us to collect more speech data via bootstrapping with the intent-based dialogue application we have built, and the hierarchy we defined will allow us to eliminate costly annotation efforts in the future, especially for the word-level slots and intent keywords. Once enough domain-specific multi-modal data is collected, our future work is to explore training end-to-end dialogue agents for our in-cabin use-cases. We are planning to exploit other modalities for improved understanding of the in-cabin dialogue as well.

\subsubsection*{Acknowledgments.} We would like to show our gratitude to our colleagues from Intel Labs, especially Cagri Tanriover for his tremendous efforts in coordinating and implementing the vehicle instrumentation to enhance multi-modal data collection setup (as he illustrated in Fig.~\ref{fig:car}), John Sherry and Richard Beckwith for their insight and expertise that greatly assisted the gathering of this UX grounded and ecologically valid dataset (via scavenger hunt protocol and WoZ research design). The authors are also immensely grateful to the members of GlobalMe, Inc., particularly Rick Lin and Sophie Salonga, for their extensive efforts in organizing and executing the data collection, transcription, and certain annotation tasks for this research in collaboration with our team at Intel Labs.

\bibliographystyle{splncs04}
%\bibliography{mlits.bib}
%\bibliography{HCII.bib}
\bibliography{CICLing.bib}

\end{document}